\documentclass[journal]{IEEEtran}
\ifCLASSINFOpdf
\else
\fi

\usepackage{amssymb}
\usepackage{lipsum}

\usepackage{graphicx}      
\usepackage{caption}
\usepackage{subcaption}    
\usepackage{dblfloatfix}   
\usepackage{placeins}      
\usepackage{amsmath}
\usepackage{booktabs}
\usepackage{multirow}
\usepackage{siunitx}
\usepackage{threeparttable}
\usepackage{makecell}
\usepackage{adjustbox}
\usepackage{tabularx}    

\newcolumntype{d}{S[table-format=3.1, table-text-post=\%, table-space-text-post=\%]}

\usepackage{caption}
\begin{document}

\title{A Context-Aware Temporal Modeling through Unified Multi-Scale Temporal Encoding and Hierarchical Sequence Learning for Single-Channel EEG Sleep Staging}

\author{Amirali~Vakili\textsuperscript{a,1}, Salar~Jahanshiri\textsuperscript{a,1}, Armin~Salimi-Badr,~\textit{Senior Member, IEEE}\textsuperscript{a,*}}

\markboth{}%
{Vakili \MakeLowercase{\textit{et al.}}: Context-Aware Temporal Modeling for Single-Channel EEG Sleep Staging}

\maketitle

\begingroup
\renewcommand\thefootnote{}\footnotetext{\textsuperscript{a}Faculty of Computer Science and Engineering, Shahid Beheshti University, Tehran, Iran.}
\renewcommand\thefootnote{*}\footnotetext{Corresponding author}
\renewcommand\thefootnote{}\footnotetext{\textit{Email address}: a\_salimibadr@sbu.ac.ir (Armin Salimi-Badr)}
\renewcommand\thefootnote{1}\footnotetext{These authors contributed equally to this work.}
\endgroup
\setcounter{footnote}{0}

\begin{abstract}
Automatic sleep staging is a critical task in healthcare due to the global prevalence of sleep disorders. This study focuses on single-channel electroencephalography (EEG), a practical and widely available signal for automatic sleep staging. Existing approaches face challenges such as class imbalance, limited receptive-field modeling, and insufficient interpretability. This work proposes a context-aware and interpretable framework for single-channel EEG sleep staging, with particular emphasis on improving detection of the N1 stage. Many prior models operate as black boxes with stacked layers, lacking clearly defined and interpretable feature extraction roles. The proposed model combines compact multi-scale feature extraction with temporal modeling to capture both local and long-range dependencies. To address data imbalance, especially in the N1 stage, class-weighted loss functions and data augmentation are applied. EEG signals are segmented into sub-epoch chunks, and final predictions are obtained by averaging softmax probabilities across chunks, enhancing contextual representation and robustness. The proposed framework achieves an overall accuracy of 89.72\% and a macro-average F1-score of 85.46\%. Notably, it attains an F1-score of 61.7\% for the challenging N1 stage, demonstrating a substantial improvement over previous methods on the SleepEDF datasets. These results indicate that the proposed approach effectively improves sleep staging performance while maintaining interpretability and suitability for real-world clinical applications.
\end{abstract}


%
\IEEEpeerreviewmaketitle

\section{Introduction}
\frenchspacing
\label{introduction}
Artificial intelligence (AI) has become a transformative tool in modern disease diagnosis, enabling automated, data-driven analysis across a wide range of medical applications, including medical imaging, biomedical signal processing, and clinical decision support systems \cite{bakator2018deep,ahsan2022machine}. By learning complex patterns from large-scale and high-dimensional healthcare data, AI-based methods have demonstrated the potential to improve diagnostic accuracy, reduce clinician workload, and enable early detection of diseases. In particular, deep learning techniques have shown remarkable success in modeling temporal and spatial dependencies in physiological signals, making them well suited for tasks involving continuous health monitoring \cite{10599257,li2014deep,liu2014early}.

Sleep plays an important role in maintaining and regulating the body's biological functions at the molecular level. Proper sleep restores human physical and mental health and helps ensure optimal performance throughout the day \cite{luyster2012sleep,medic2017short}.
Sleep is divided into three main types: non-rapid eye movement sleep (NREM), rapid eye movement sleep (REM), and the wake state. NREM sleep itself includes three stages: N1, N2, and N3, with the last stage transitioning into REM sleep \cite{wolpert1969manual}. These two types of sleep alternate cyclically during the sleep process, and if this cycle is disrupted or certain stages are skipped, sleep quality is impaired.

In today's world, the assessment of sleep stages is mainly performed manually by specialists in this field, which comes with limitations. Sleep experts classify sleep stages using hypnograms; however, they may struggle to accurately detect slow changes in brain signals or interpret complex rules. Furthermore, differences of opinion among experts can reduce accuracy \cite{danker2009interrater,rosenberg2013american}.

Other challenges also exist, such as patient comfort, diagnostic costs, and the use of portable and wearable devices. To address these issues, we need a fully automated and low-cost method for performing this task.

Early attempts at automation were carried out using classical machine learning algorithms. Those approaches relied heavily on handcrafted features extracted from different physiological signals such as EEG, EOG, and EMG \cite{gunecs2010efficient,lajnef2015learning,10824785,10595067,11012662}. Typical classifiers used in these studies include SVM, Random Forest, K-NN, Decision Trees, and Ensemble Methods. For example, Willemen et al. \cite{6600727} employed statistical features alongside an SVM classifier, reaching 69\% accuracy. Research has also focused on developing efficient methods using a single sensor to enhance practicality. Dimitriadis et al. \cite{DIMITRIADIS2018815}   proposed a single-EEG-sensor technique based on the dynamic reconfiguration of cross-frequency coupling (CFC) using a Naive Bayes classifier. Similarly, Zhu et al. \cite{6733276} introduced a novel approach by mapping single-channel EEG signals into difference visibility graphs (DVG) to extract graph-based features, which were then classified with an SVM, achieving an accuracy of 87.5\% for six sleep stages.

As a replacement for classical methods, deep learning models have been used to switch from handcrafted features to representation learning. Eldele et al. \cite{9417097} proposed a novel attention-based deep learning architecture called AttnSleep to classify sleep stages using single-channel EEG signals. Their model begins with a feature extraction module based on a multi-resolution convolutional neural network (MRCNN) and adaptive feature recalibration (AFR), followed by a temporal context encoder (TCE) that leverages a multi-head attention mechanism with causal convolutions to capture temporal dependencies. Evaluations on three public datasets demonstrated that AttnSleep outperformed state-of-the-art techniques in terms of different evaluation metrics.

Another method was proposed by Supratak et al. \cite{7961240}, who developed DeepSleepNet, a deep learning model for automatic sleep stage scoring based on raw single-channel EEG. The model combines convolutional neural networks for time-invariant feature extraction and bidirectional long short-term memory (BiLSTM) networks to learn transition rules among sleep stages. Tested on multiple datasets with different sampling rates and scoring standards, DeepSleepNet achieved overall accuracies of 86.2\% (MASS) and 82.0\% (SleepEDF).

Phan et al. \cite{phan2019seqsleepnet} introduced SeqSleepNet, a hierarchical recurrent neural network designed to perform sequence-to-sequence sleep staging. At the epoch level, the network employs a filterbank layer to learn frequency-domain filters and an attention-based recurrent layer for short-term sequential modeling, while at the sequence level, another recurrent layer models long-term dependencies across multiple epochs. Their method achieved an overall accuracy of 87.1\% and a macro F1-score of 83.3\% on a large public dataset .

Mousavi et al. \cite{mousavi2019sleepeegnet} presented SleepEEGNet, a sequence-to-sequence deep learning approach using single-channel EEG signals. The model integrates deep convolutional neural networks to extract temporal and frequency features with a recurrent architecture to capture long-term dependencies. To mitigate class imbalance, novel loss functions were applied to balance misclassification errors across stages. On the SleepEDF dataset, SleepEEGNet achieved 84.26\% overall accuracy and a macro F1-score of 79.66\%, outperforming many existing methods.

The main problem in this field is the lack of samples in the N1 class, and N1 samples are mainly misclassified as Wake or N2. Different methods have been used to resolve this issue. Xuewei Cheng \cite{cheng2024sleepegan} used GANs to augment data and increase the number of samples for minority classes, achieving an accuracy of 86.8\%. Many other researchers used weighted cross-entropy to tackle this problem. Although this approach increases accuracy in the N1 class, it leads to lower accuracy in other classes, making it an inappropriate solution.
In this paper, we address these challenges through a context-aware preprocessing strategy and a novel unified architecture. Our main contributions are:
\begin{itemize}
    \item A novel preprocessing strategy with sub-epoch segmentation and voting to address limited receptive field modeling by enhancing context and increasing training samples
    \item A custom CNN architecture with multi-scale kernels and squeeze-and-excitation blocks for interpretable feature extraction with defined layer roles, enabling channel attention and temporal compression
    \item A hierarchical BiLSTM with attention pooling to capture both short-term and long-term temporal dependencies, addressing the need for better temporal modeling
    \item Comprehensive data augmentation strategies and class-weighted loss functions to mitigate the severe class imbalance problem, particularly for the challenging N1 stage
\end{itemize}

The remainder of this paper is organized as follows: Section~\ref{sec:methods} describes our proposed methodology, including data preprocessing, temporal feature extraction, and sequence modeling. Section~\ref{sec:expandres} presents our experimental setup and results, including comparisons with prior methods and ablation studies. Section~\ref{sec:dis} discusses the implications of our findings, and Section~\ref{sec:conc} concludes the paper.

\begin{figure*}[t]
    \centering
    \includegraphics[width=\textwidth]{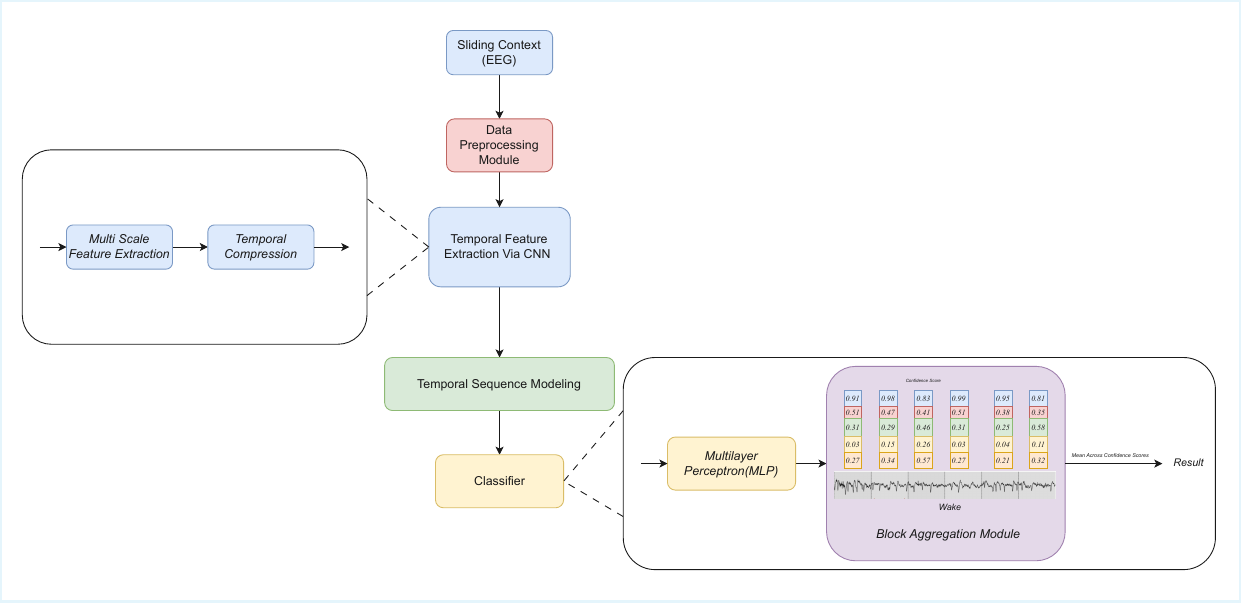}
    \caption{Overall overview of the proposed pipeline.}
    \label{fig:overview}
\end{figure*}


\begin{figure*}[h]
    \centering
    \includegraphics[width=\textwidth]{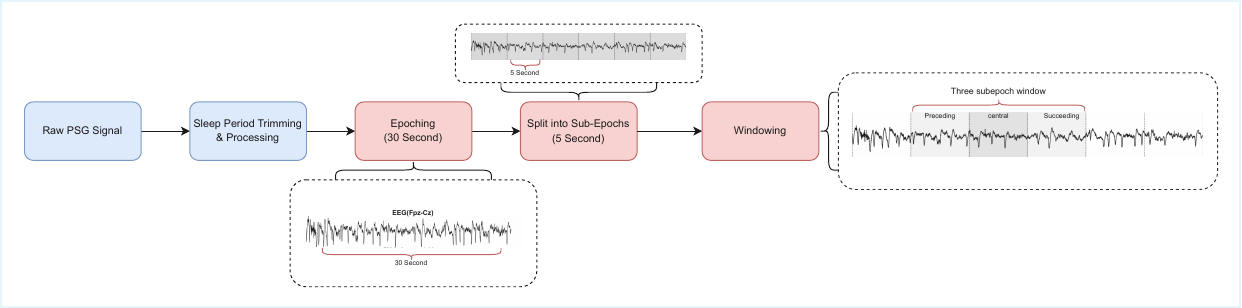}
    \caption{The structure of the preprocessing pipeline.}
    \label{fig:preproc}
\end{figure*}

\section{Proposed Methods}
\label{sec:methods}

Fig.~\ref{fig:overview} presents an overview of our proposed pipeline. The methodology consists of four main stages: (1) data preprocessing with sub-epoch segmentation and context window construction, (2) temporal feature extraction via a custom CNN with multi-scale kernels and squeeze-and-excitation blocks, (3) hierarchical temporal sequence modeling using BiLSTM with attention pooling, and (4) classification with data augmentation and class-weighted loss functions to address class imbalance. Each component is designed to address specific challenges in automatic sleep staging, as detailed in the following subsections.

\subsection{Data Preprocessing}
We have employed some novel preprocessing methods in this area alongside previously used ones from earlier works \cite{8502139}. As shown in Fig.~\ref{fig:preproc}, trimmed signals were segmented into 30-second epochs. In addition, each of these epochs was divided into 6 non-overlapping 5-second sub-epochs to have high-resolution signals for subsequent analysis. Furthermore, the model has to be evaluated using 30-second epochs to be comparable with other papers, so we assigned each 5-second chunk a block identifier that is relevant to its parent. Thereby, in the classification module, predictions can be mapped back to the original epoch from which the sub-epochs originated.

Instead of looking at each 5-second piece of data individually, a three-sub-epoch context window was applied to each central sub-epoch, and was grouped with the piece before it and the piece after it. This approach captures short-term temporal dependencies, and the contextual information surrounding the central segment can be used. For each three-sub-epoch window, the central sub-epoch’s label was assigned as the target.

\begin{figure*}[t]
    \centering
    \includegraphics[width=\textwidth,keepaspectratio]{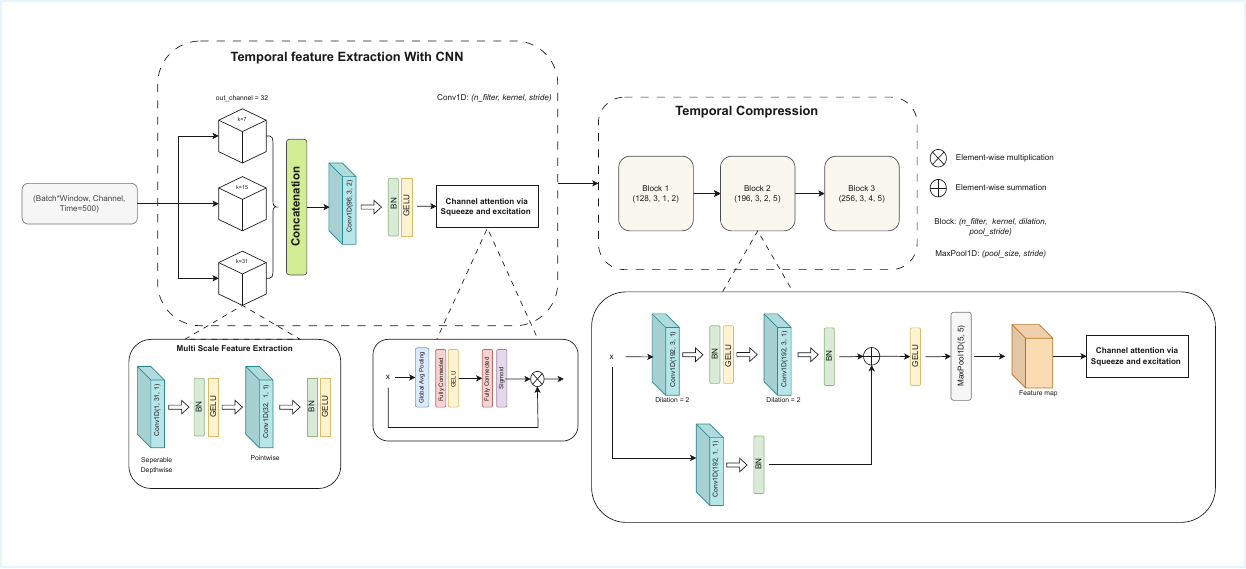}
    \caption{CNN-based temporal feature extraction and compression.}
    \label{fig:conv}
\end{figure*}

\subsection{Temporal Feature Extraction via CNN}
\label{sec:tfe-cnn}
The overall architecture of the temporal feature extraction module, including
the multi-scale convolutional branches and temporal compression pathway, is illustrated in Fig.~\ref{fig:conv}.
After the preprocessing stage, the data are passed to the \emph{Temporal Feature Extraction via CNN} component.
It is composed of two sub-blocks, namely \emph{Multi-Scale Feature Extraction} and \emph{Temporal Compression},
which are described in Sections~\ref{sec:multiscale} and \ref{sec:tempcomp}.

\emph{All convolutions in this module are 1D, operating along the temporal dimension.}

The input and output tensors are defined as:
\begin{equation}
\mathbf{X} \in \mathbb{R}^{B \cdot W \times C_{in} \times T_{in}},
\quad
\mathbf{Y} \in \mathbb{R}^{B \cdot W \times C_{out} \times T_{out}} \, ,
\end{equation}
where $B$ is the batch size, $W$ the number of windows, $C_{in}=1$ the number of input channels,
and $T_{in}=500$ the number of samples per window.
The output tensor has $C_{out}=256$ channels and $T_{out}=5$ latent time steps.

At the end of this component, the original 500-sample window (5 s at 100 Hz)
is reduced to 5 latent time steps, corresponding to a temporal compression factor:
\begin{equation}
T_{out} = \frac{T_{in}}{100} \, .
\end{equation}

Band-specific information (e.g., delta 0.5--4 Hz, theta 4--8 Hz, spindle 11--16 Hz) is first extracted by the convolutional branches prior to temporal downsampling; the latent sequence summarizes these learned features across time.

\subsubsection{Multi-Scale Feature Extraction}
\label{sec:multiscale}

The first sub-block is the \emph{Multi-Scale Feature Extraction} stage,
designed to capture multi-scale temporal features across different frequency ranges.
It employs three parallel convolutional branches with kernel sizes of
7, 15, and 31 samples, respectively, targeting short-, medium-,
and long-term dependencies.

Each branch applies depthwise separable convolution \cite{howard2017mobilenets} followed by a $1 \times 1$
pointwise convolution. The parameter counts are given by:
\begin{equation}
\text{Params}_{\text{full}} = k \cdot C_{in} \cdot C_{out},
\quad
\text{Params}_{\text{dw}} = k \cdot C_{in} + C_{in} \cdot C_{out} \, ,
\end{equation}
where $k$ denotes the kernel size.
This design significantly reduces computation compared to full convolution.

The outputs of the three branches are concatenated to form a unified feature map:
\begin{equation}
\mathbf{F} = \text{Concat}\big(\mathbf{F}_{k=7}, \mathbf{F}_{k=15}, \mathbf{F}_{k=31}\big) \, .
\end{equation}
Following concatenation, a Gaussian Error Linear Unit (GELU) \cite{hendrycks2016gaussian} nonlinearity is applied,
and a $3 \times 1$ convolution with stride 2 is used for early temporal pooling.

A squeeze-and-excitation (SE) \cite{hu2018squeeze} block is then employed to enhance channel-wise feature quality:
\begin{equation}
\mathbf{s} = \sigma \!\left( W_2 \, \delta \,\big(W_1 \cdot \text{GAP}(\mathbf{U})\big) \right),
\quad
\mathbf{X}' = \mathbf{s} \odot \mathbf{U} \, ,
\end{equation}
where $\text{GAP}$ denotes global average pooling, $\delta$ is the GELU activation,
$\sigma$ is the sigmoid function, and $r=8$ is the reduction ratio.

The Multi-Scale Feature Extraction stage contains approximately $30.6 \times 10^3$ parameters,
most of which are concentrated in the reduction convolution stage.
Overall, it achieves efficient multi-frequency receptive field fusion
while maintaining a modest computational budget.

\subsubsection{Temporal Compression}
\label{sec:tempcomp}

The second stage, \emph{Temporal Compression}, is designed to reduce temporal resolution and expand feature channels, while capturing long-range dependencies with dilated convolutions \cite{yu2015multi}. Each of the three residual blocks contains two 1D convolutions, batch normalization, GELU activation, a residual (or projected) shortcut, max-pooling for downsampling, and a squeeze–and–excitation (SE) module for channel reweighting.
The dilated 1D convolution used in this stage is defined as:
\begin{equation}
y_o[t] = \sum_{i=1}^{C_{\text{in}}} \sum_{m=0}^{k-1}
W_{o,i}[m] \; x_i[t - d \cdot m] \, .
\end{equation}

Each residual block applies two such convolutions, followed by normalization and activation. The residual unit with projection is given by:
\begin{align}
\mathbf{z}_1 &= \phi \Big( \mathrm{BN}_1 \big( \mathrm{Conv}_{k_1,d_1}(\mathbf{x}) \big) \Big) , \\[8pt]
\mathbf{z}_2 &= \mathrm{BN}_2 \big( \mathrm{Conv}_{k_2,d_2}(\mathbf{z}_1) \big) , \\[8pt]
\mathbf{y}   &= \phi \Big( \mathbf{z}_2 + \mathcal{P}(\mathbf{x}) \Big) .
\end{align}

\begin{figure*}[t]
    \centering
    \includegraphics[width=\textwidth,keepaspectratio]{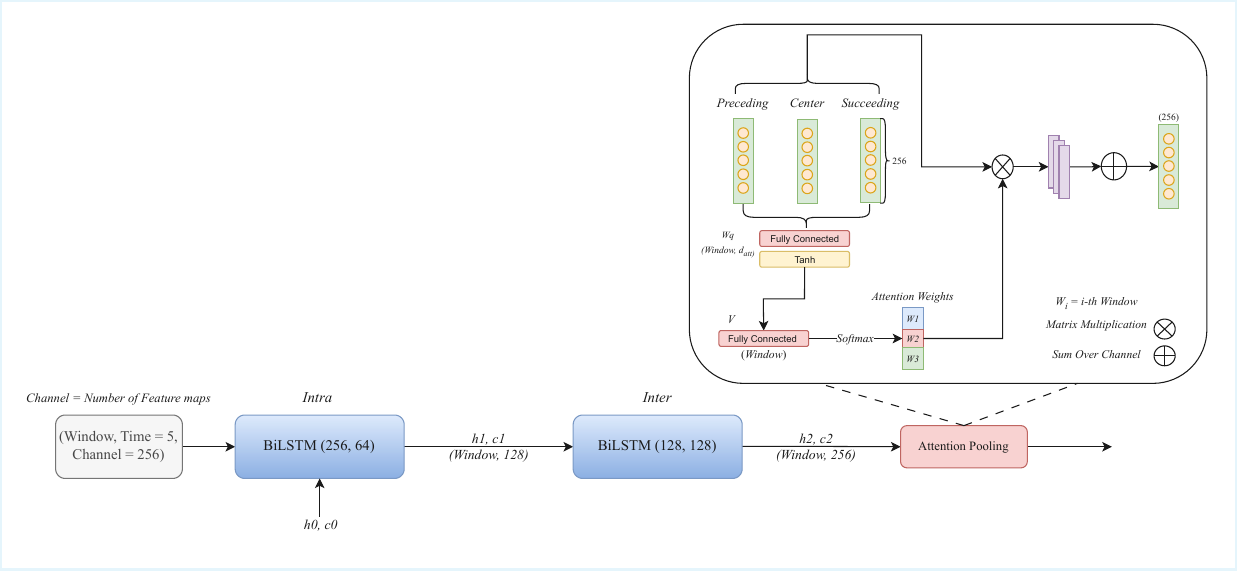}
    \caption{Hierarchical BiLSTM-based temporal sequence modeling.}
    \label{fig:sequence}
\end{figure*}

After the residual operation, temporal downsampling is performed using max pooling:
\begin{equation}
\tilde{\mathbf{y}} = \mathrm{MaxPool}_{s_i}(\mathbf{y}),
\qquad
T_i = \left\lfloor \frac{T_{i-1}}{s_i} \right\rfloor ,
\end{equation}
We use three temporal strides, specifically $(s_1, s_2, s_3) = (2, 5, 5)$.
Applying these strides reduces the sequence length step by step:
starting from $T_0 = 250$, it becomes $T_1 = 125$, then $T_2 = 25$,
and finally $T_3 = 5$.

Across the convolutional blocks, the number of channels increases
progressively. The model starts with $C_0 = 96$ channels in the input
stage, then expands to $128$ in block 1, $192$ in block 2, and
reaches $256$ channels in block 3.

Overall, the input sequence (which is $500$ samples after preprocessing)
is compressed down to just $5$ latent steps. This corresponds to a
temporal compression factor of $500$ divided by $5$, which equals $100$.

Because dilations $(1,2,4)$ are used across blocks, the receptive field grows to cover a broader temporal context. The receptive field increment at block $i$ is:
\begin{equation}
\Delta \mathrm{RF}_i = (k_1-1)\,d_1 + (k_2-1)\,d_2 ,
\end{equation}
and accumulates as:
\begin{align}
\mathrm{RF}_i &= \mathrm{RF}_{i-1}
+ \Big[(k_1-1)\,d_1 + (k_2-1)\,d_2\Big] \cdot
\prod_{j=0}^{i-1} s_j , \\[6pt]
s_0 &= 2 \;\; (\text{multi-scale stride}) .
\end{align}

Finally, this stage outputs:
\begin{equation}
\mathbf{Y}_{\text{res}} \in \mathbb{R}^{B \cdot W \times 256 \times 5} \, ,
\end{equation}
which is passed to the subsequent LSTM head for higher-level temporal modeling.

\subsection{Temporal Sequence Modeling}

Fig.~\ref{fig:sequence} illustrates the hierarchical temporal sequence modeling
block, which integrates an intra-window BiLSTM, an inter-window BiLSTM
\cite{chung2016hierarchical}, and an additive attention pooling mechanism
\cite{chorowski2015attention}. This block operates after the CNN feature extractor
and is responsible for modeling both short- and long-range temporal dependencies.
The input to this component (output of the previous module) is

\begin{equation}
\mathbf{Z} \in \mathbb{R}^{B \cdot W \times F \times 5},
\end{equation}
where \(F=256\) and the final dimension corresponds to the 5 latent timesteps per window. Here, \(B\) denotes the batch size and \(W\) denotes the number of context windows (set to \(W=3\) in our experiments). The temporal modeling stage yields a fixed-size embedding \(\mathbf{s} \in \mathbb{R}^{2H_2}\) (here \(2H_2=256\)), which is used as input to the MLP classifier.

\subsubsection{Intra-window BiLSTM}
This module operates within each 5-step window. First, the input is permuted to match the format expected by PyTorch LSTMs:
\begin{equation}
\mathbf{Z} \;\rightarrow\; \mathbb{R}^{B \cdot W \times 5 \times F}.
\end{equation}
A bidirectional LSTM with hidden size \(H_1=64\) per direction is applied across the 5 latent timesteps. We take the last timestep output to obtain a per-window representation \(\tilde{\mathbf{h}}_i \in \mathbb{R}^{2H_1}\). These representations capture fine-grained micro-temporal dynamics within each 5-step latent sequence, preserving short-range dependencies learned during compression.

\subsubsection{Inter-window BiLSTM}
The per-window representations are stacked back into a sequence ordered as past, center, and future windows:
\begin{equation}
\tilde{\mathbf{H}} \in \mathbb{R}^{B \times W \times 2H_1}.
\end{equation}
This sequence forms the input to the inter-window BiLSTM.
A second bidirectional LSTM is applied across the \(W\) windows with hidden size \(H_2=128\) per direction, producing
\begin{equation}
\mathbf{U} \in \mathbb{R}^{B \times W \times 2H_2}.
\end{equation}
This module models relationships between neighboring sub-epochs (sleep micro-architecture).

\begin{table*}[htbp]
\centering
\caption{Summary of employed datasets.}
\small
\setlength{\tabcolsep}{7.5pt} 
\renewcommand{\arraystretch}{1.6} 
\begin{tabular}{|c|c|c|c|c|c|c|c|}
\hline
Dataset & EEG channel & \multicolumn{6}{c|}{Epoch counts and proportion} \\
\cline{3-8}
 &  & Wake & N1 & N2 & N3 & REM & Total \\
\hline
SleepEDF-20 & Fpz--Cz &
8285 (19.6\%) &
2804 (6.6\%) &
17799 (42.1\%) &
5703 (13.5\%) &
7717 (18.2\%) &
42308 \\
\hline
SleepEDF-78 & Fpz--Cz &
69824 (35.0\%) &
21522 (10.8\%) &
69132 (34.7\%) &
13039 (6.5\%) &
25835 (13.0\%) &
199352 \\
\hline
\end{tabular}
\label{Table1}
\end{table*}

\subsubsection{Additive Attention Pooling}
Given \(\mathbf{U}\), we pool information across the \(W\) windows into a single vector. For each window \(t\),
\begin{equation}
\mathbf{s}_t = \tanh\!\big(W_q \mathbf{U}_t\big), \qquad e_t = \mathbf{v}^\top \mathbf{s}_t ,
\end{equation}
where \(W_q \in \mathbb{R}^{2H_2 \times d_{\text{att}}}\) and \(\mathbf{v} \in \mathbb{R}^{d_{\text{att}}}\). Attention weights are obtained via a softmax across windows:
\begin{equation}
\alpha_t = \frac{\exp(e_t)}{\sum_{\tau=1}^{W} \exp(e_{\tau})}.
\end{equation}
The sequence is then aggregated by a weighted sum, yielding the fixed-length embedding
\begin{equation}
\mathbf{s} = \sum_{t=1}^{W} \alpha_t \, \mathbf{U}_t \;\in\; \mathbb{R}^{2H_2}.
\end{equation}
With \(d_{\text{att}}=64\), the attention weights have shape \(\boldsymbol{\alpha} \in \mathbb{R}^{B \times W}\), and the pooled vector has shape \(\mathbf{s} \in \mathbb{R}^{2H_2}\). Each weight in \(\boldsymbol{\alpha}\) indicates the contribution of a sub-epoch to the final inference and is retained for analysis. The attention parameters \((W_q,\mathbf{v})\) are learned jointly with the rest of the network; no auxiliary supervision is used. Finally, \(\mathbf{s}\) is passed to the MLP classifier.

\subsection{MLP Classifier}

After the LSTM layer, the features extracted by that layer are passed to a final classification module. This module is primarily employed to transform the high-level temporal features into the final sleep stage probabilities. The Hierarchical LSTM head produces a final feature vector of 256 dimensions (2 $\times$ 128 for the bidirectional hidden states) for each input sequence. Before this module, a dropout and normalization layer is used to prevent overfitting and to stabilize the learning process by normalizing the features across the entire feature dimension for each sample. These normalized features are passed to an MLP for the final classification.

The proposed MLP consists of two hidden layers and a final output layer. The first hidden layer is a linear layer that maps the 256-dimensional input to 512 neurons, followed by batch normalization, GELU, and dropout. In the second hidden layer, we employ a linear layer that maps 512 to 256 neurons, again followed by batch normalization, GELU, and dropout.

In the final output layer, a linear layer maps the 256-dimensional input to 5 neurons. Each of these neurons represents one of the Wake, N1, N2, N3, or REM classes. The output of this MLP is a vector of logits for each sleep stage, from which the final sleep stage prediction is derived.

\subsection{Loss Function}

We employed weighted cross-entropy for this multi-class classification problem. We used this approach because it is a standard loss function for multi-class classification, as it measures the dissimilarity between predicted probability distributions and true labels.

The main issue with sleep datasets is that they suffer from severe class imbalance. To address this, we employed weighted cross-entropy, where each class is weighted inversely proportional to its frequency in the training set. Additionally, we applied weight decay (L2 regularization) to the model parameters in order to reduce overfitting and improve generalization.

The weighted loss with weight decay is computed as:
\begin{equation}
\mathcal{L}
= - \sum_{b=1}^{B} \sum_{k=1}^{K} w_k \, y_{b,k} \, \log \hat{p}_{b,k}
\;+\;
\lambda \sum_{j} \left\lVert \theta_j \right\rVert_2^2 ,
\end{equation}
where $w_k$ is the weight for class $k$, $y_{b,k}$ is the ground-truth one-hot label for sample $b$, $\hat{p}_{b,k}$ is the predicted probability for class $k$, $\theta_j$ represents the trainable model parameters, and $\lambda$ is the weight decay coefficient.

\begin{table}[htbp]
\centering
\caption{Experimental Hyperparameters.}
\label{tab:hyperparameters}
\small  
\setlength{\tabcolsep}{6pt} 
\begin{tabular}{|l|l|}
\hline
Hyperparameter & Value \\
\hline
Learning Rate (lr) & 1e-4 \\
Weight Decay & 1e-4 \\
Batch Size & 128 \\
Max Epochs & 100 \\
Early Stopping Patience & 30 \\
Learning Rate Scheduler & ReduceLROnPlateau \\
Decay Factor of Scheduler & 0.5\\
Patience of Scheduler & 7 epochs\\
\hline
\end{tabular}
\end{table}

\section{Experiments and Results}
\label{sec:expandres}

To evaluate our proposed method, we performed a series of experiments to analyze it from several aspects. First, we examined our overall performance under a cross-validation protocol and analyze the outcomes across different stages. Secondly, we compared the overall and per-class performance of our method with established baselines using standard evaluation metrics to show its distinctive advantages. Finally, we performed an ablation study to show the contribution of individual components of our method to the overall performance.

\begin{table*}[htbp]
\centering
\caption{Per-class performance on SleepEDF-20 and SleepEDF-78 using the Fpz--Cz EEG channel.}
\label{tab:per_class_performance_combined}
\small
\setlength{\tabcolsep}{8pt}
\renewcommand{\arraystretch}{1.2}
\begin{tabular}{|c|c|c|c|c|c|c|}
\hline
Dataset & Class & F1 & Precision & Sensitivity & Specificity & AUROC \\
\hline
\multirow{5}{*}{SleepEDF-20}
 & WAKE & 94.80 & 95.70 & 93.90 & 99.00 & 99.70 \\
 & N1   & 61.70 & 68.90 & 56.00 & 98.20 & 95.90 \\
 & N2   & 91.40 & 90.80 & 91.90 & 93.30 & 98.00 \\
 & N3   & 89.70 & 89.70 & 89.70 & 98.40 & 99.40 \\
 & REM  & 89.80 & 87.30 & 92.40 & 97.00 & 99.10 \\
\hline
\multirow{5}{*}{SleepEDF-78}
 & WAKE & 95.00 & 96.00 & 94.00 & 98.60 & 99.46 \\
 & N1   & 59.00 & 64.00 & 54.00 & 97.00 & 93.29 \\
 & N2   & 87.00 & 86.00 & 89.00 & 94.00 & 97.06 \\
 & N3   & 83.00 & 83.00 & 84.00 & 98.00 & 99.28 \\
 & REM  & 85.00 & 85.00 & 88.00 & 96.00 & 98.72 \\
\hline
\end{tabular}
\end{table*}

\begin{table*}[htbp]
\centering
\footnotesize
\setlength{\tabcolsep}{8pt}
\renewcommand{\arraystretch}{1.2}
\caption{Per-class comparison of F1, Precision, and Sensitivity on SleepEDF-20 and SleepEDF-78 using the Fpz--Cz EEG channel. Best results are highlighted in bold.}

\label{tab:perclass}
\begin{tabular}{|c|c|ccc|ccc|}
\hline
\multirow{2}{*}{Model} &
\multirow{2}{*}{Stage} &
\multicolumn{3}{c|}{SleepEDF-20} &
\multicolumn{3}{c|}{SleepEDF-78} \\
\cline{3-8}
 &  & F1 & Precision & Sensitivity &
 F1 & Precision & Sensitivity \\
\hline

\multirow{5}{*}{Ours}
 & WAKE & \textbf{94.8} & \textbf{95.7} & \textbf{93.9} &
 \textbf{95.0} & \textbf{96.0} & \textbf{94.0} \\
 & N1   & \textbf{61.7} & \textbf{68.9} & \textbf{56.0} &
 \textbf{59.0} & \textbf{64.0} & \textbf{54.0} \\
 & N2   & \textbf{91.4} & 90.8 & \textbf{91.9} &
 \textbf{87.0} & 86.0 & \textbf{89.0} \\
 & N3   & 89.7 & 89.7 & 89.7 &
 83.0 & 83.0 & 84.0 \\
 & REM  & \textbf{89.8} & 87.3 & \textbf{92.4} &
 \textbf{85.0} & 85.0 & \textbf{88.0} \\
\hline

\multirow{5}{*}{AttnSleep~\cite{9417097}}
 & WAKE & 89.7 & 89.6 & 89.7 & 92.0 & 92.3 & 91.8 \\
 & N1   & 42.8 & 47.1 & 39.1 & 42.1 & 45.3 & 39.2 \\
 & N2   & 88.8 & 89.1 & 88.6 & 85.0 & 83.5 & 86.5 \\
 & N3   & \textbf{90.2} & \textbf{90.7} & 89.8 & \textbf{82.1} & \textbf{82.3} & 82.0 \\
 & REM  & 79.0 & 76.1 & 82.2 & 74.2 & 73.1 & 75.3 \\
\hline

\multirow{5}{*}{SleepEEGNet~\cite{mousavi2019sleepeegnet}}
 & WAKE & 89.1 & 87.8 & 90.5 & 91.7 & -- & -- \\
 & N1   & 52.1 & 50.0 & 54.5 & 44.0 & -- & -- \\
 & N2   & 86.7 & \textbf{91.2} & 82.7 & 82.4 & -- & --\\
 & N3   & 85.1 & 81.6 & 88.8 & 73.6 & -- & -- \\
 & REM  & 85.0 & 81.6 & 88.7 & 76.0 & -- & -- \\
\hline

\multirow{5}{*}{U-Time~\cite{perslev2019u}}
 & WAKE & 87.0 & -- & -- & 92.0 & -- & -- \\
 & N1   & 52.0 & -- & -- & 51.0 & -- & -- \\
 & N2   & 86.0 & -- & -- & 84.0 & -- & -- \\
 & N3   & 84.0 & -- & -- & 75.0 & -- & -- \\
 & REM  & 84.0 & -- & -- & 80.0 & -- & -- \\
\hline

\multirow{5}{*}{DeepSleepNet\cite{7961240}}
 & WAKE & 84.7 & 86.0 & 83.4 & -- & -- & -- \\
 & N1   & 46.6 & 43.5 & 50.1 & -- & -- & -- \\
 & N2   & 85.9 & 90.5 & 81.7 & -- & -- & -- \\
 & N3   & 84.8 & 77.1 & \textbf{94.2} & -- & -- & -- \\
 & REM  & 82.4 & 80.9 & 83.9 & -- & -- & -- \\
\hline

\multirow{5}{*}{SleepEGAN~\cite{cheng2024sleepegan}}
 & WAKE & 91.7 & -- & -- & 93.1 & -- & -- \\
 & N1   & 53.6 & -- & -- & 51.7 & -- & -- \\
 & N2   & 89.2 & -- & -- & 85.8 & -- & -- \\
 & N3   & 89.1 & -- & -- & 81.2 & -- & -- \\
 & REM  & 86.1 & -- & -- & 82.0 & -- & -- \\
\hline

\multirow{5}{*}{SleepFocalNet~\cite{zan2025temporal}}
 & WAKE & 93.4 & 94.0 & 92.8 & 93.7 & 92.9 & 94.6 \\
 & N1   & 55.4 & 61.4 & 50.5 & 52.3 & 60.4 & 46.2 \\
 & N2   & 90.4 & 89.1 & 91.7 & 86.9 & 84.9 & 89.1 \\
 & N3   & 89.5 & 89.4 & 89.6 & 80.1 & 84.2 & 76.5 \\
 & REM  & 89.1 & 88.8 & 89.5 & 87.1 & 84.5 & 89.9 \\
\hline

\end{tabular}
\end{table*}

\subsection{Dataset}

In this study, we evaluated our method using two benchmark datasets from the SleepEDF Database Expanded: SleepEDF-20 and SleepEDF-78. SleepEDF-20 consists of recordings of 20 subjects between 25 and 34 years of age, a total of 39 recordings. Each PSG record contains multiple channels such as EEG (Fpz-Cz and Pz-Oz), horizontal EOG (ROC-LOC), and chin EMG, sampled at 100 Hz. Following the Rechtschaffen and Kales (R\&K) standard, recordings were labeled as Wake (W), stages N1–N4, rapid eye movement (REM), Movement, or Unknown \cite{kemp2000analysis}. According to common and standard practice, we excluded the Movement and Unknown epochs and merged N3 and N4 into a single N3 stage, producing the standard five-class classification: [Wake (W), N1, N2, N3, REM]. SleepEDF-78 is a later expanded version of SleepEDF. It includes 197 polysomnography (PSG) recordings from 78 healthy participants (34 male and 44 female) aged between 25 and 101 years. For most subjects, two consecutive day–night recordings were available, except for three cases (subjects 13, 36, and 52). Similar to SleepEDF-20, PSG records in SleepEDF-78 contain EEG (Fpz–Cz and Pz–Oz), horizontal EOG (ROC–LOC), and chin EMG signals, sampled at 100 Hz and annotated with hypnograms in 30-second epochs by expert specialists. A summary of the dataset statistics is provided in Table~\ref{Table1}. In this paper, we used the Fpz–Cz EEG channel  as inputs for the experimental analysis.

\subsection{Data Augmentation}

Various augmentation methods were applied to increase model stability and generalization, as well as mitigating overfitting problem. Each method was applied stochastically to simulate realistic variability and artifacts. The methods included:
\begin{itemize}
    \item Additive Gaussian Noise: For simulating electrode and environmental noise \cite{bishop1995training}.
    \item Temporal Scaling and Shifting: randomly compressing and stretching the signal to simulate physiological variability.
    \item Temporal Masking: zeroing out short intervals to force the model to exploit contextual information. This technique is also used to augment minority class samples.
\end{itemize}

\subsection{Experimental Setups}

To evaluate our proposed model, we performed 5-fold cross-validation using Stratified Group K-Fold cross-validation to ensure subject independence and prevent data leakage by grouping epochs based on block identifiers according to our preprocessing method. The model was trained using the Adam optimizer along with a learning rate scheduler and weight decay for regularization. Early stopping was applied after 30 epochs without improvement in validation accuracy. As mentioned previously, Due to the imbalanced dataset, the model was trained with weighted cross-entropy loss due to the imbalanced dataset. All experiments were performed on an NVIDIA A100 GPU via the Google Colab Pro Service. The hyperparameters are detailed in Table~\ref{tab:hyperparameters}.

\subsection{Evaluation Metrics}

Overall performance of our model was evaluated using metrics commonly employed in previous studies, including accuracy (Acc.), macro-average F1-score (MF1), Cohen’s kappa coefficient (K), sensitivity (Sens.), and specificity (Spec.). Additionally, for per-class analysis, we used per-class F1-score ($F_1$), per-class precision ($P_{\text{rec}}$), per-class sensitivity ($P_{\text{sens}}$), and per-class area under the receiver operating characteristic curve (AUROC). We also presented detailed results for each class using the confusion matrix.

\subsection{Overall Cross-validated Performance}

\begin{figure}[!t]
  \centering
  \includegraphics[width=\columnwidth,keepaspectratio]{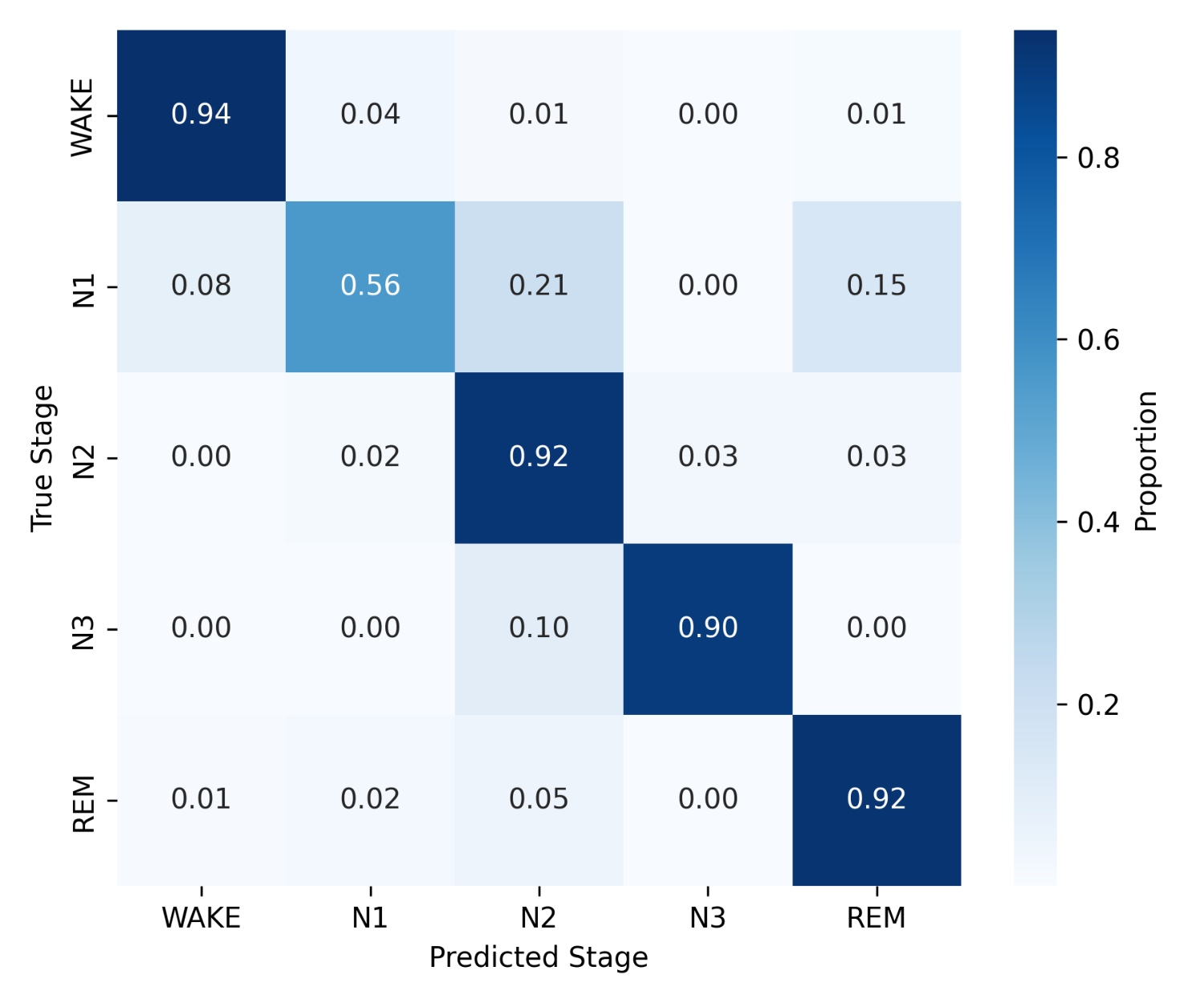}
  \caption{Confusion matrix from 5-fold cross-validation on the Sleep-EDF20 dataset.}
  \label{fig:cv_confmat}
\end{figure}

\begin{figure*}[!t]
  \centering
  \includegraphics[width=\textwidth,keepaspectratio]{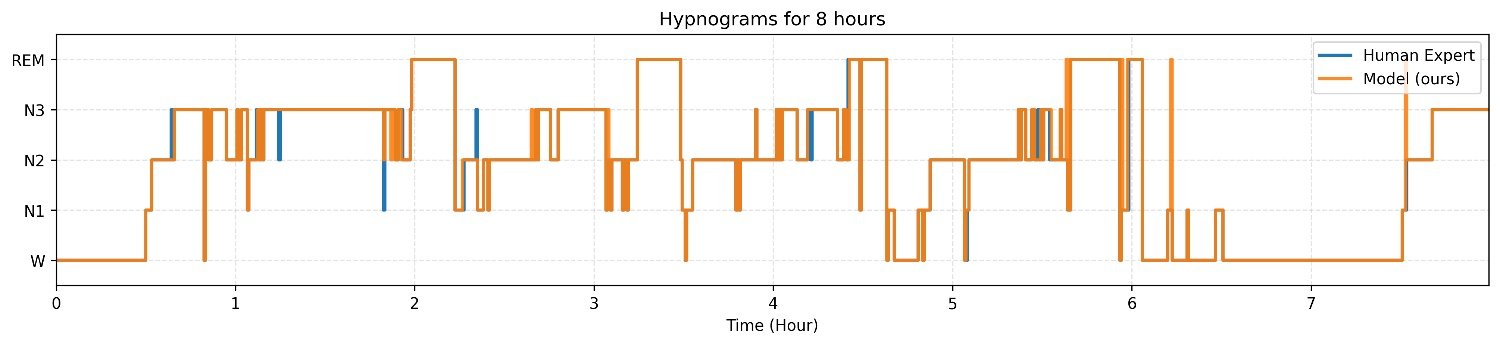}
  \caption{Hypnograms over 8 hours comparing human expert annotations to our model predictions.}
  \label{fig:hypnogram}
\end{figure*}

Overall performance of the 5-fold cross-validation was detailed in Tables~\ref{tab:per_class_performance_combined}, ~\ref{tab:perclass}, and ~\ref{tab:overall}.
The corresponding learning curves, illustrating the evolution of training and
validation accuracy across epochs, are shown in Fig.~\ref{fig:cv_curves}.
The results indicate that the proposed model achieves stable performance with
low variance across folds. As illustrated in the confusion matrix
(Fig.~\ref{fig:cv_confmat}), the proposed model achieves high confidence and reliability
in Wake, N2, N3, and REM stages along with N1. However, most misclassifications
occur in the N1 stage, with misclassifications to N2, Wake, and REM. A comparison
between expert-annotated and model-predicted hypnograms is shown in
Fig.~\ref{fig:hypnogram}. Our method is compared with the following baselines:

\begin{figure}[!t]
  \centering
  \includegraphics[width=\columnwidth,keepaspectratio]{cv_learning_curves_accuracy.png}
  \caption{Cross-validation learning curves showing mean training and validation accuracy per epoch with $\pm1\sigma$ shading across folds.}
  \label{fig:cv_curves}
\end{figure}

\begin{table*}[htbp]
\centering
\footnotesize
\setlength{\tabcolsep}{8pt} 
\renewcommand{\arraystretch}{1.25} 
\caption{Overall performance comparison on SleepEDF-20 and SleepEDF-78 using the Fpz--Cz EEG channel. Best and second-best results are highlighted in bold and underlined, respectively.}
\label{tab:overall}
\begin{tabular}{|c|cccccc|cccccc|}
\hline
\multirow{2}{*}{Model} &
\multicolumn{6}{c|}{SleepEDF-20} &
\multicolumn{6}{c|}{SleepEDF-78} \\
\cline{2-13}
 & Acc. & Kappa & Macro F1 & Sens. & Spec. &  &
 Acc. & Kappa & Macro F1 & Sens. & Spec. &  \\
\hline

Ours          & \textbf{89.72} & \textbf{85.85} & \textbf{85.46} & \textbf{84.78} & \textbf{97.16} &  & \textbf{86.41} & \textbf{81.25} & \textbf{81.97} & \textbf{81.86} & \textbf{96.39} &  \\
SeqSleepNet~\cite{phan2019seqsleepnet}            & 82.20 & 79.80 & 78.40 & 78.00 & 96.10 &  & 82.60 & 76.00 & 76.40 & 76.30 & 95.40 &  \\
AttnSleep~\cite{9417097}             & 84.40 & 79.00 & 78.10 & -- & -- &  & 81.30 & 74.30 & 75.10 & -- & -- &  \\
SleepEEGNet~\cite{mousavi2019sleepeegnet}            & 84.26 & 79.00 & 79.66 & -- & -- &  & 80.03 & 73.00 & 73.55 & -- & -- &  \\
U-Time~\cite{perslev2019u}                 & 82.10 & 78.00 & 78.80 & -- & -- &  & 82.10 & 73.00 & 76.50 & -- & -- &  \\
DeepSleepNet~\cite{7961240}          & 80.80 & 73.10 & 74.20 & -- & -- &  & 78.50 & 70.20 & 75.30 & 75.00 & 95.10 &  \\
XSleepNet~\cite{phan2021xsleepnet}              & 86.30 & 81.30 & 80.60 & 80.20 & 96.40 &  & 84.00 & 77.80 & 77.90 & \underline{77.50} & 95.70 &  \\
SleepGCN~\cite{wang2024sleepgcn}               & 86.96 & 81.68 & 81.81 & \underline{82.85} & \underline{96.54} &  & 83.79 & 77.62 & 76.19 & 75.53 & \underline{95.67} &  \\
SleepEGAN~\cite{cheng2024sleepegan}              & 86.80 & 82.00 & 81.90 & -- & -- &  & 83.80 & \underline{82.00} & 78.70 & -- & -- &  \\
SleepFocalNet~\cite{zan2025temporal}          & \underline{88.50} & \underline{84.10} & \underline{83.60} & -- & -- &  & \underline{85.50} & 80.00 & \underline{79.80} & -- & -- &  \\
\hline
\end{tabular}
\end{table*}

\begin{enumerate}
    \item DeepSleepNet used CNN layers to extract time-invariant features from raw single-channel EEG, followed by BiLSTM layers to model temporal dependencies between epochs. The model was trained in two stages (pretraining and fine-tuning) to capture both short- and long-term sleep patterns \cite{7961240}. \\
    \item SeqSleepNet reformulated the sleep staging task as a sequence-to-sequence classification problem. It used parallel filterbanks and BiRNN layers to extract time-frequency features \cite{phan2019seqsleepnet} . \\
    \item AttnSleep used multi-scale CNN branches with attention to capture multi-scale EEG features while emphasizing salient patterns \cite{9417097}. \\
    \item U-Time used a fully convolutional U-Net followed by an LSTM layer to capture temporal dependencies \cite{perslev2019u}. \\
    \item SleepEEGNet combined CNNs and BiRNNs to extract time-frequency domain features and capture temporal dependencies, respectively \cite{mousavi2019sleepeegnet}. \\
    \item SleepEGAN introduced a generative adversarial network (GAN) framework for sleep staging to augment minority sleep stages, such as N1 \cite{cheng2024sleepegan}.\\
    \item SleepGCN introduced a graph convolutional framework to explicitly model sleep transition rules, combining a residual network with an LSTM branch for feature extraction \cite{wang2024sleepgcn}. \\
    \item SleepFocalNet used focal modulation with multi-scale feature extraction via CNNs \cite{zan2025temporal}. \\
    \item XSleepNet leveraged BiRNNs to fuse raw signal data with time–frequency image features, creating a joint representation to support sleep stage prediction \cite{phan2021xsleepnet}.\\
\end{enumerate}

\begin{table*}[htbp]
\centering
\caption{Ablation study on SleepEDF-20.}
\small
\renewcommand{\arraystretch}{1.25}
\setlength{\tabcolsep}{6pt}

\begin{tabular}{|p{7.0cm}|c|c|c|c|c|c|}
\hline
Methods &
Acc. &
Macro F1 &
Kappa &
Sens. &
Spec. &
N1-Class F1 \\
\hline

Multi-Scale Feature Extraction Only &
84.39 & 80.60 & 79.16 & 84.03 & 96.21 & 53.50 \\
\hline

Multi-Scale Feature Extraction + Temporal Compression &
86.13 & 79.85 & 80.30 & 76.37 & 95.68 & 57.00 \\
\hline

Multi-Scale Feature Extraction + Temporal Compression + Temporal Sequence Modeling &
88.53 & 83.60 & 84.08 & 80.90 & 96.10 & 59.00 \\
\hline

Multi-Scale Feature Extraction + Temporal Compression + Temporal Sequence Modeling + Augmentation &
\textbf{89.72} & \textbf{85.46} & \textbf{85.85} & \textbf{84.78} & \textbf{97.16} & \textbf{61.70} \\
\hline

\end{tabular}
\label{table:ablation}
\end{table*}

As summarized in Table~\ref{tab:overall}, our proposed method outperforms all baselines in terms of every metric on the SleepEDF-20 dataset, achieving the highest accuracy (89.72\%), kappa (85.85), macro-F1 (85.46\%), sensitivity (84.78\%), and specificity (97.16\%). A similar performance trend is observed on the SleepEDF-78 dataset, as reported in Table~\ref{tab:overall}. As shown in the per-class analysis in Table~\ref{tab:perclass}, our method achieves the highest F1-score and sensitivity in most sleep stages, with only limited exceptions where AttnSleep, DeepSleepNet, SleepEEGNet, or SleepFocalNet perform marginally better in a single metric. Remarkably, our method demonstrates significant improvement in the challenging N1 stage, achieving F1-scores of 61.7\% and 59.0\% on the SleepEDF-20 and SleepEDF-78 datasets, respectively. Overall, these results demonstrate that our proposed method consistently outperforms existing single-channel EEG methods in both global and class-level performance.

\begin{figure*}[t]
    \centering
    \includegraphics[width=\textwidth]{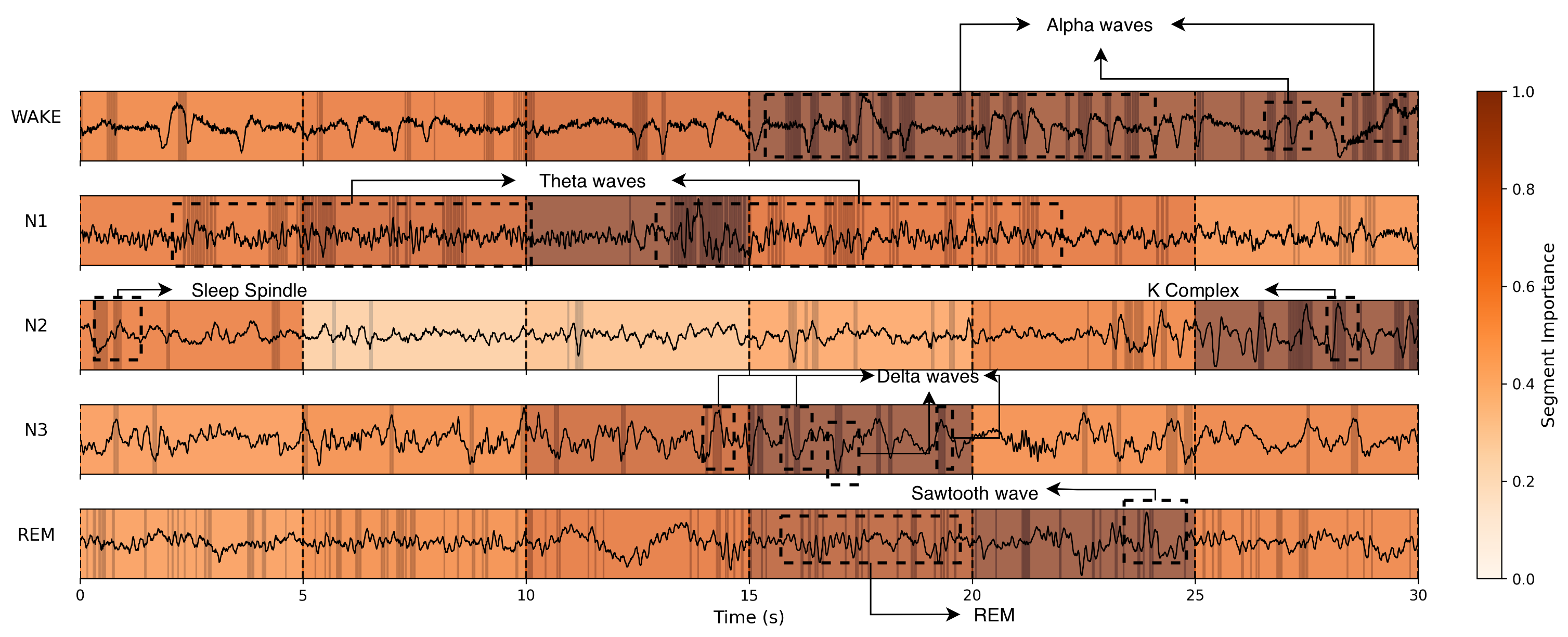}
    \caption{Segment-importance visualization for representative EEG segments across sleep stages. Darker regions indicate higher importance, highlighting characteristic physiological patterns captured by our model. Dotted boxes mark representative time intervals corresponding to salient stage-specific patterns.}
    \label{fig:segment_importance}
\end{figure*}

\begin{figure}[h]
    \centering
    \begin{subfigure}{0.48\columnwidth}
        \centering
        \includegraphics[width=\linewidth]{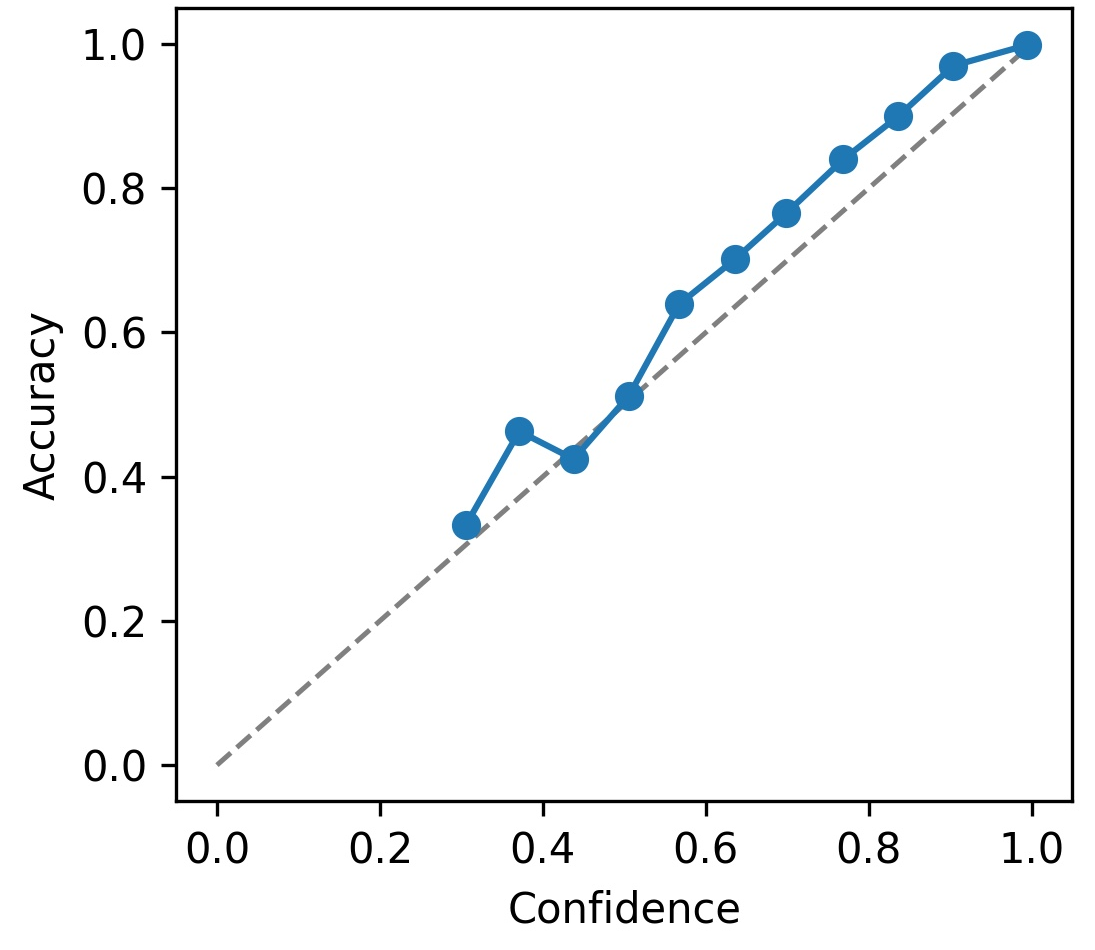}
        \caption{Overall reliability}
    \end{subfigure}
    \hfill
    \begin{subfigure}{0.48\columnwidth}
        \centering
        \includegraphics[width=\linewidth]{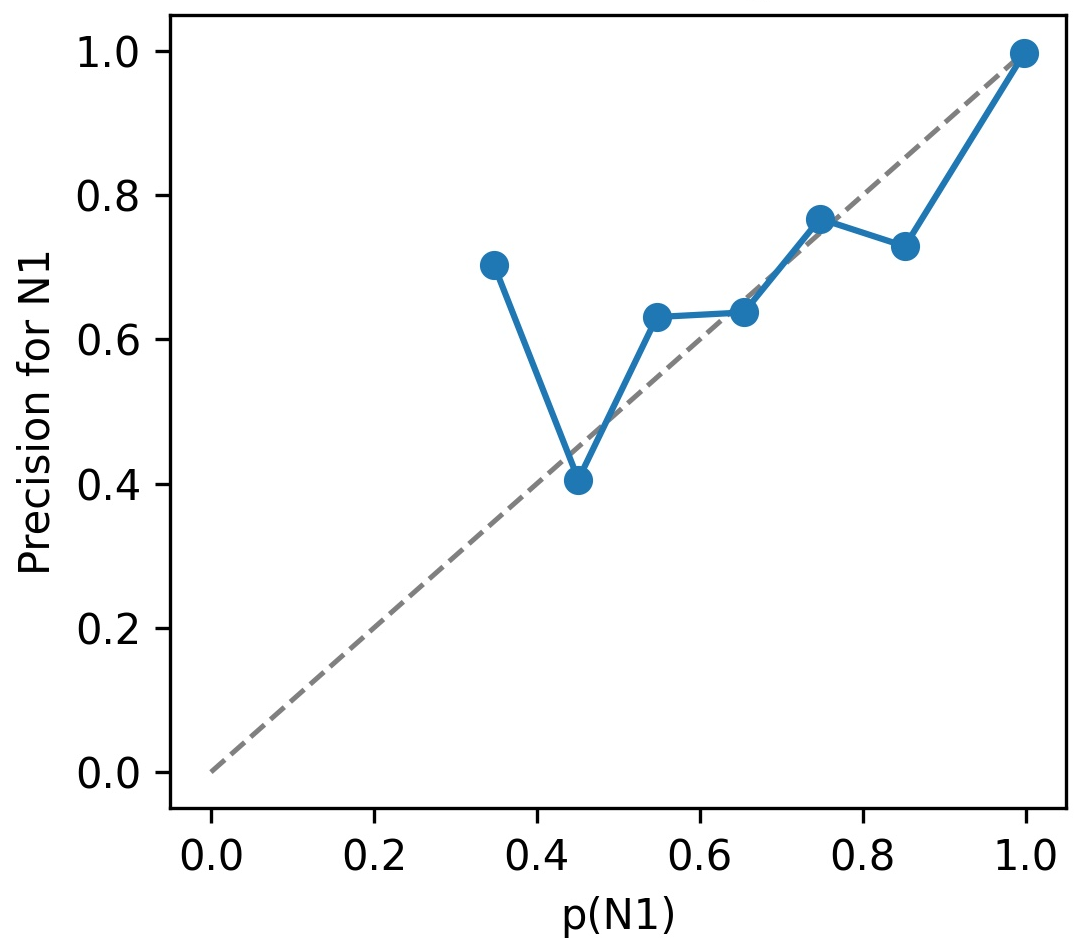}
        \caption{N1 reliability}
    \end{subfigure}
    \caption{Reliability diagrams for overall predictions and the N1 sleep stage.}
    \label{fig:reliability}
\end{figure}

\subsection{Ablation Study}
\label{sec:ablation}
This section presents the ablation study applied to examine the contribution of each proposed architecture using the SleepEDF-20 dataset. As illustrated in Table~\ref{table:ablation}, a clear view of each component's contribution is provided, including multi-scale convolution blocks, temporal compression, BiLSTM-based sequence modeling, and ultimately data augmentation. As a baseline model, the multi-scale feature extraction module was examined and achieved 84.39\% accuracy and 80.60\% macro-F1 score. These results show that using multi-scale kernels improves performance by extracting features at different temporal resolutions. Adding the temporal compression module significantly increased accuracy to 86.13\%. Temporal downsampling represented each signal with compact feature map that retains rich information while expanding model's receptive field. Further enhancement is achieved with the integration of the hierarchical temporal modeling block, which consists of intra-window BiLSTM, inter-window BiLSTM, and additive attention mechanisms. This resulted in 88.53\% accuracy with 83.60\% macro-F1 score. The results highlight that capturing contextual relationships is particularly important for distinguishing transitional and ambiguous sleep stages, such as N1 and REM. Finally, by adding data augmentation techniques, the complete model reached the highest performance level with 89.72\% accuracy and a macro-F1 score of 85.46\%. The provided data augmentation techniques significantly enhanced the model’s ability to generalize, particularly for the N1 stage, which resulted in a 2.7\% increase in the N1 class-wise F1 score. While temporal compression and hierarchical sequence modeling accounted for the most substantial improvements, data augmentation added the robustness necessary to achieve the best performance across all evaluated metrics. Overall, the ablation study demonstrates that each component makes a meaningful contribution to the proposed model’s final performance.

\section{Discussion}
\label{sec:dis}

The findings show that the proposed method effectively mitigates key challenges in automatic single-channel EEG sleep staging. In particular, it addresses limited receptive-field modeling and the persistent class-imbalance problem, especially for the N1 stage. The proposed method achieves consistent improvements compared with existing single-channel baselines. The preprocessing module is a key contributor to these gains, as it enables the model to capture short-term transitions with finer temporal resolution. Moreover, the windowing strategy allows the model to incorporate contextual information that is often lost at the conventional 30-second epoch level. As illustrated in Fig.~\ref{fig:segment_importance}, the model primarily focuses on established physiological EEG patterns, which enhances the interpretability and clinical relevance of its predictions. Specifically, alpha activity is emphasized during wakefulness, theta waves during N1, spindles and K-complexes during N2, delta waves during N3, and mixed low-voltage activity with sawtooth waves during the REM stage. Fig.~\ref{fig:cv_curves} further shows that the learning curves converge steadily with minimal variance, indicating that the model generalizes well. As shown in the confusion matrix (Fig.~\ref{fig:cv_confmat}) and the per-class results (Table~\ref{tab:per_class_performance_combined}), the model performs reliably in predicting the Wake, N2, N3, and REM stages. In addition, its performance on the N1 stage surpasses that of existing single-channel EEG baselines. Except for N1, the remaining stages exhibit more distinctive and stable patterns, which enables the model to achieve high sensitivity and specificity. Nevertheless, N1 remains the most challenging stage to classify, consistent with previous studies. This difficulty arises from overlapping patterns with neighboring stages, particularly wakefulness, as well as the inherently transitional nature of N1. Despite these challenges, the proposed method achieves an F1-score of 61.7\% on the N1 stage using single-channel EEG, representing a substantial improvement over baseline performance. This improvement is attributed to the use of sub-epoch segmentation in combination with enhanced architectural modules, class-weighted loss, and effective data augmentation strategies, which collectively improve the training distribution of minority-stage samples. Furthermore, Fig.~\ref{fig:reliability} shows that, even for the challenging N1 stage, the predicted confidence scores are relatively well calibrated. The overall reliability diagram in Fig.~\ref{fig:reliability} further demonstrates good calibration across all sleep stages, with a low Expected Calibration Error of 0.010. By integrating multi-scale feature extraction, temporal compression via CNNs, hierarchical sequence modeling, and data augmentation within a class-weighted loss framework, the proposed approach achieves strong state-of-the-art-level performance on global evaluation metrics. In addition, the interpretability offered by segment-importance maps and calibration analysis highlights the potential of this framework for semi-automated clinical workflows. Overall, the proposed method provides principled and effective solutions to key challenges in automatic sleep staging and consistently outperforms related models, as demonstrated in Tables~\ref{tab:per_class_performance_combined}and ~\ref{tab:overall}.

\section{Summary and Conclusions}
\label{sec:conc}
As discussed earlier, data imbalance and limited receptive-field modeling remain among the most critical challenges in automatic sleep staging. In addition, developing lightweight models that can operate efficiently in real-time settings is an important practical requirement. In this work, we proposed a novel and lightweight single-channel EEG-based architecture built upon an efficient multi-branch convolutional compressor and a hierarchical BiLSTM framework, combined with class-weighted loss functions and data augmentation strategies to address class imbalance. Using only a single EEG channel, the proposed model achieved an accuracy of 89.72\% and a macro-average F1-score of 85.46\% under 5-fold cross-validation on the SleepEDF-20 dataset. Notably, the model attained an F1-score of 61.7\% for the challenging N1 stage, representing a substantial improvement over existing single-channel EEG-based approaches. Overall, the proposed method outperformed prior techniques while maintaining computational efficiency and interpretability. Future work may focus on further improving model reliability and adaptability in practical scenarios. Moreover, evaluating the approach on clinical datasets and wearable devices would provide deeper insight into its performance under real-world conditions.

\FloatBarrier

\bibliographystyle{ieeetr}
\bibliography{mybib}

\end{document}